\documentclass[letterpaper, 10 pt, conference]{ieeeconf}  

\usepackage{cite}
\usepackage{graphicx}

\IEEEoverridecommandlockouts                              

\title{\LARGE \bf
Control of a Back-Support Exoskeleton to Assist Carrying Activities
}

\author{Maria Lazzaroni$^{1}$, Giorgia Chini$^{2}$, Francesco Draicchio$^{2}$, Christian Di Natali$^{1}$, Darwin G. Caldwell$^{1}$ \\ and Jes{\'u}s Ortiz$^{1}$ 
\thanks{This work is funded by the Italian Workers' Compensation Authority INAIL.}
\thanks{$^1$Department of Advanced Robotics, Istituto Italiano di Tecnologia (IIT), Genova, Italy (corresponding author: {\tt\small maria.lazzaroni@iit.it}).}
\thanks{$^2$Department of Occupational and Environmental Medicine, Epidemiology and Hygiene, INAIL, Rome, Italy.}
}

\begin{document}

\maketitle
\thispagestyle{empty}
\pagestyle{empty}

\begin{abstract}
Back-support exoskeletons are commonly used in the workplace to reduce low back pain risk for workers performing demanding activities. 
However, for the assistance of tasks differing from lifting, back-support exoskeletons potential has not been exploited extensively.

This work focuses on the use of an active back-support exoskeleton to assist carrying. 
A control strategy is designed that modulates the exoskeleton torques to comply with the task assistance requirements. 
In particular, two gait phase detection frameworks are exploited to adapt the exoskeleton assistance according to the legs' motion. 
The control strategy is assessed through an experimental analysis on ten subjects. 
Carrying task is performed without and with the exoskeleton assistance.
Results prove the potential of the presented control in assisting the task without hindering the gait movement and improving the usability experienced by users. 
Moreover, the exoskeleton assistance significantly reduces the lumbar load associated with the task, demonstrating its promising use for risk mitigation in the workplace.


\end{abstract}

\section{INTRODUCTION}
%
Despite the ongoing trend in automation in industry, various work tasks are still difficult to automate due to their complexity or prohibitive cost \cite{Frohm2008_AutomationManufacturing} \cite{Nitsche2021_AutomationLogistics}. 
For instance, in many sectors spanning from automotive and manufacturing to logistics and construction, several demanding activities are still performed manually by workers.
Manual material handling (MMH) activities such as assembling tasks, pushing and pulling, lifting and carrying of heavy objects significantly load workers' musculoskeletal systems, increasing the risk of developing musculoskeletal disorders (MSDs) \cite{Kumar2001_JointOverloading} \cite{Coenen2004_LumbarCompression}. 
Work-related MSDs are the most common occupational disease in many industrialized countries \cite{Punnett2004_MSDworld}, with high costs to enterprises and society and impacts on workers' quality of life. 
From the latest report \cite{deKok2019_MSDimpactEurope}, roughly three out of every five workers in the EU-28 report MSDs complaints. 
The most common MSD type reported by workers is low back pain (LBP), which only in Europe generates an annual cost of about 240 billion euros (2\% of GDP) \cite{Bevan2012_LBPimpactEurope}.

To mitigate the risk of work-related LBP, occupational back-support exoskeletons are effective and high-potential solutions that are being introduced in the workplace \cite{Baldassarre2022_ExoskeletonsWorkpalce}. Back-support exoskeletons limit workers' lumbar load by reducing the activation of the back muscles and, thus, the muscles' contribution to lumbar compression \cite{Toxiri2019_ReviewExoskeletons} \cite{DeLooze2016_ReviewExoskeletons}. 
Both passive and active exoskeletons have reported beneficial effects when supporting forward bending and symmetric lifting \cite{Kermavnar2020_ExoskeletonEffectsReview}.
However, for the execution of other MMH tasks in the workplace, the use of back-support exoskeletons has not been explored extensively. 
Indeed, the ISO 11228 and NIOSH standards have classified carrying, pulling and pushing tasks as risky as lifting \cite{Cheung2007_NIOSHguidelinesMMH}. 
Likewise, from epidemiological studies, it emerges that the execution of carrying exposes workers to an increased risk of developing LBP \cite{Marras2000_OccupationalLBP}.

In the literature, previous works exist that employ exoskeletons for walking assistance, supporting the users' hip flexion-extension or ankle plantarflexion.
These devices, however, were designed to aid individuals with compromised abilities who cannot produce adequate forces for locomotion \cite{Panizzolo2016_ExosuitWalkingMetabolic} \cite{Lenzi2013_BackSupportExoskeletonWalking}.
Specifically regarding occupational back-support exoskeletons, few studies have evaluated their performance during walking or carrying. 
As expected, passive exoskeletons, which provide symmetric assistance, are perceived by users as restricting the legs' natural movement during walking \cite{Naf2018_SpexorEffect, Baltrusch2019_LAEVOEvaluationLiftingWalking, Kozinc2021_SpexorEvaluationWalking, Baltrusch2018_LaevoEffect}. 
Other studies have evaluated active exoskeletons in assisting carrying \cite{Poliero2020_XoTrunkCarrying} \cite{Wei2020_MeBotEXO}. 
However, in these cases, the assistance is not tailored for the specific task but was designed for lifting, thus it is symmetric and constrains the users' natural movement.
While there is a benefit for the back muscles, there is a clear hindrance for the legs: hip and knee Range of Motion (RoM) is reduced, stride duration increases, and speed decreases \cite{Poliero2020_XoTrunkCarrying}.

This work presents the use of a back-support exoskeleton for assisting carrying tasks to reduce the risk associated with this task in the workplace.
We designed and tested a control strategy specifically tailored for assisting carrying on an active exoskeleton. 
The strategy is completely described in Section \ref{Sec:Exoskeleton_control}.
The details of the experimental evaluation (experimental design and assessed metrics) are reported in Section \ref{sec:EXPERIMENTALEVALUATION}.
The experimental data are presented in Section \ref{sec:results}. 
In Section \ref{sec:Discussion}, the results are discussed, along with the main improvements obtained by the new control.
Finally, conclusions are drawn in Section \ref{sec:conclusions}.

\section{EXOSKELETON CONTROL}
\label{Sec:Exoskeleton_control}
\begin{figure*}[thpb]
	\centering
        \includegraphics[scale=0.76]{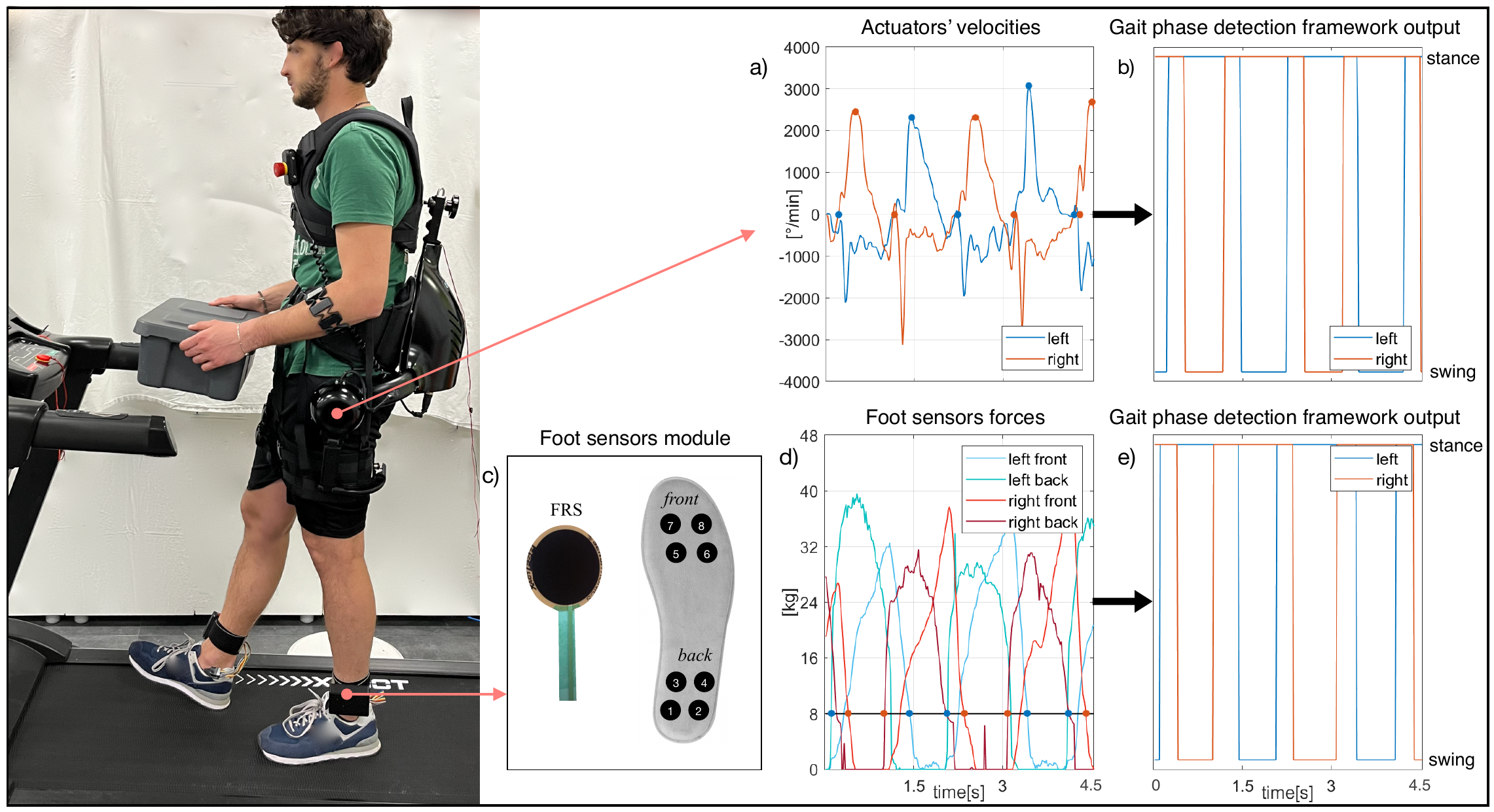}
	\caption{\label{fig_setup_withXoTrunk} Experimental set-up with a subject performing the test with the exoskeleton. The two gait phase detection frameworks are shown. The \textit{foot sensors} framework is based on 8 FSRs placed under the plantar area of each foot: 4 sensors on the \textit{front} and 4 sensors on the \textit{back} (c). The signals from the two clusters (\textit{front} and \textit{back}) of the two feet are displayed for one subject (d), for example, over two strides (4,5 seconds). The legs' gait phases, i.e., the gait detection framework output computed based on the foot sensors forces, are displayed on the right (e). The \textit{actuators velocity} framework is based on the angular velocity of the exoskeleton actuators and is presented for the same subject over the same time interval (a). In the figure, the points that define the start and end of the gait phases are indicated (i.e., for each leg, the moment when the contralateral angular velocity crosses zero and the moment when the ipsilateral angular velocity reaches the peak). The gait phases of each leg are the framework output (b).}
\end{figure*}
Our study focuses on carrying task, given its relevance for LBP risk in the workplace. 
We design a control strategy with the aim to improve the effectiveness of the exoskeleton and overcome the limitations of the current devices in assisting this specific task: in particular, the hindrance and interference with the legs.
We propose two frameworks that perform gait phase detection during walking with the exoskeleton, named from the sensors used to identify the legs' gait phases: 1) \textit{foot sensors} framework and 2) \textit{actuators velocity} framework.
Based on gait phases, the strategy controls the exoskeleton assistance by modulating the torques (100 Hz).

\subsection{Gait phase detection frameworks}
\label{SubSec:Gait_phase_detection_frameworks}
The first gait phase detection framework is based on force sensors placed on the plantar area of each foot 
\cite{Mateos2016_ExoshoeRobomate, Rana2009_FSRgait, Wang2023_GaitPhaseDetectionFramework}.
Indeed, during the gait cycle, the contacts of the foot with the ground define the leg's phase. 
Stance begins with heel strike and ends with toe-off. 
Conversely, swing begins with toe-off and ends with heel strike.
We use Force Sensing Resistors (FSRs) that exhibit varying resistance in response to force applied to the sensing area. As the force on a sensor increases, the resistance decreases.
For each foot, we integrate 8 FSRs on an insole that can be inserted into regular shoes.
The 8 sensors are divided into two clusters: 4 sensors on the \textit{front}, which cover the metatarsal area, and 4 sensors on the \textit{back}, which cover the heel area (Fig. \ref{fig_setup_withXoTrunk}). 
In this way, we can detect the correct foot pressure during the gait cycle, as most of the forces are received in the metatarsal and heel areas \cite{Rana2009_FSRgait}.

The second gait phase detection framework is based on the angular velocity of the user's hips.
For each leg, the stance phase begins when the angular velocity of the contralateral hip crosses zero, and the swing phase begins when the angular velocity of the ipsilateral hip reaches the peak value \cite{Mentiplay2018_HipAngularVelocityWalking}.
When wearing the exoskeleton, the angular velocity of the user's hips is estimated by measuring the angular velocity of the actuators with embedded encoders (Fig. \ref{fig_setup_withXoTrunk}).

\subsection{Exoskeleton control strategy}
The control strategy is implemented based on the gait phase detection frameworks.
Different from previous works, this control strategy assists the user's legs asymmetrically and independently.
The total torque $\tau_{exo}$ provided by the exoskeleton is divided between the two legs according to their gait phases, as classified by the detection frameworks:
\begin{equation}
    \tau_{stance} = K_{stance} \cdot \tau_{exo} 
    \quad
    \quad 
    \tau_{swing} = K_{swing} \cdot \tau_{exo} \
    \label{eq:tauRightLeft}
\end{equation}
For instance, during carrying, there are two possible states:
\begin{itemize}
    \item Double stance: the assistance is divided equally to the two legs: $K_{right} = K_{left} = K_{stance} = 0.5$.
    \item One leg is in stance phase and the other leg is in swing; the assistance is provided to the leg in stance phase, which supports the user's weight, while no assistance is provided to the leg in swing in order to not interfere with the user's movement: $K_{stance} = 0.5$ and $K_{swing} = 0$.
\end{itemize}
The total assistive torque $\tau_{exo}$ is defined according to the forearm muscle activity, recorded by surface electrodes of the Myo armband, a commercial device (Myo gesture control armband10, Thalmic Labs Inc.). Since the activation of forearm muscles increases when handling a load, the torque automatically accounts for the carried load. The muscle activity is normalized to the maximum value and multiplied by a user-selected gain $K_{myo}$:
\begin{equation}
    \tau_{exo} = K_{myo} \cdot \frac{EMG_{forearm}}{EMG_{forearm}^{MAX}}
    \label{eq:tauTotal}
\end{equation}
We tested the control strategy using two inputs, i.e., the two gait phase detection framework outputs.
The resulting controls are named from the framework used: 1) \textit{foot sensors} mode, 2) \textit{actuators velocity} mode.


\subsection{XoTrunk exoskeleton}
\begin{figure}[thpb]
	\centering
        \includegraphics[scale=0.26]{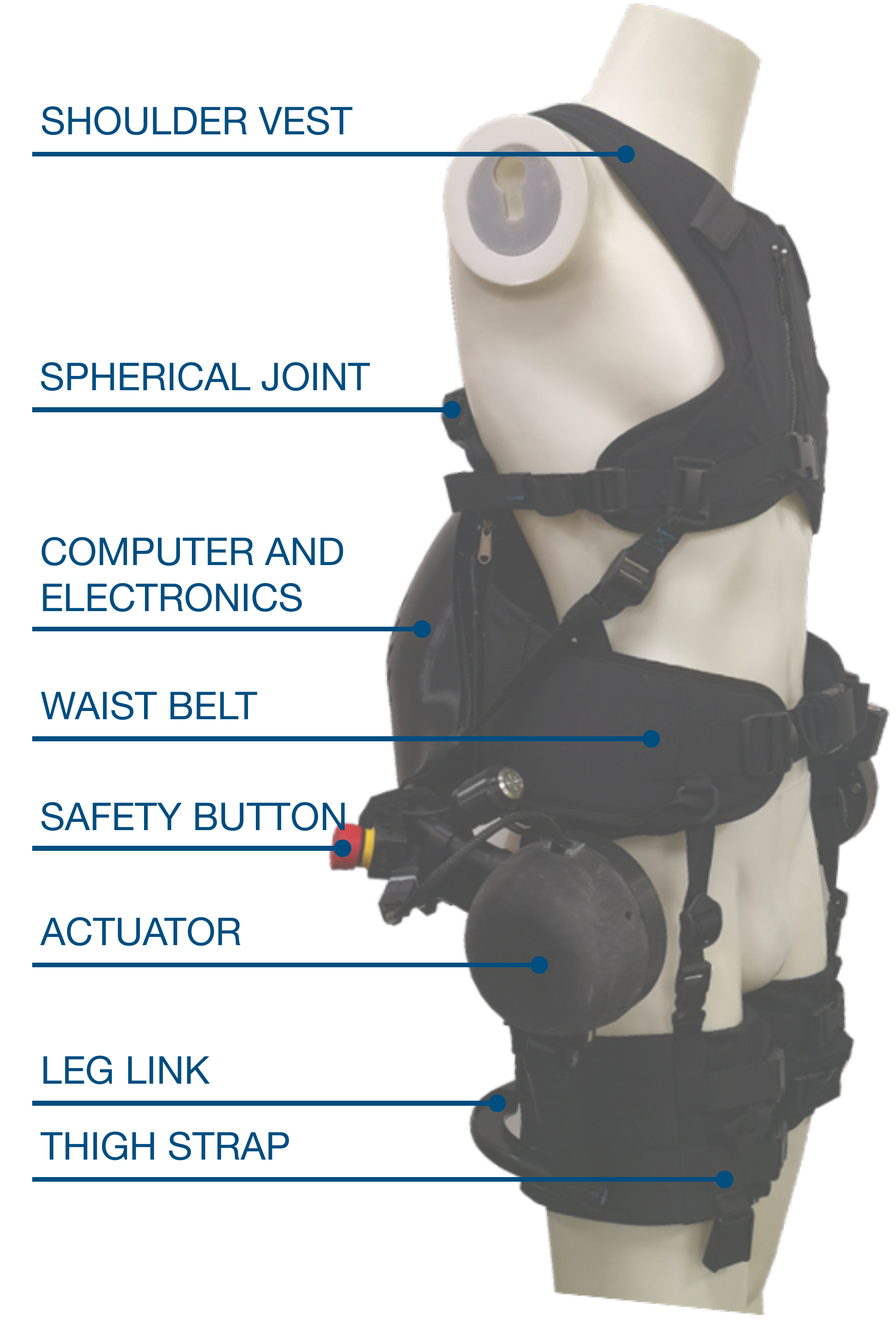}
	\caption{\label{fig_XoTrunk} The XoTrunk exoskeleton prototype.}
\end{figure}

The proposed methodology is tested using the XoTrunk exoskeleton, an active back-support exoskeleton for assisting MMH activities \cite{Lazzaroni2022_linearAcceleration}, shown in Fig. \ref{fig_XoTrunk}.
XoTrunk was developed by the XoLab at the Istituto Italiano di Tecnologia (IIT) through a collaboration with the Italian Workers’ Compensation Authority (INAIL). 
The assistance is provided by two DC brushless motors, aligned with the user's hips, that generate assistive torques in the sagittal plane.
The torque modulation is regulated by control strategies, which are selected according to the activity the user is performing to address the assistance needs.

\section{EXPERIMENTAL EVALUATION}
\label{sec:EXPERIMENTALEVALUATION}
An experimental evaluation is conducted to assess the effects of the assistance provided by the new control strategy. 
In particular, the experiment tests two hypotheses:
\begin{itemize}
    \item[] \textit{Hypothesis 1:} the exoskeleton assistance does not interfere with or hinder the legs during walking, improving the usability experienced by users with respect to previous studies.
    \item[] \textit{Hypothesis 2:} the exoskeleton assistance reduces users' lumbar load associated with the execution of carrying.
\end{itemize}

\subsection{Experimental design}
Ten male healthy subjects (age: 28.5 $\pm$ 2.7 years, weight: 75.1 $\pm$ 9.7 kg, height: \mbox{182 $\pm$ 9.0 cm}) with no history of LBP participated in the test, approved by the Ethics Committee of Liguria (reference number: CER Liguria 001/2019). 
Participants walked on a treadmill carrying a 10 kg box in three assistance conditions, performed in a randomized order:
\begin{itemize}
    \item without exoskeleton: \textit{no-exo};
    \item with \textit{exo} in the \textit{foot sensors} control mode;
    \item with \textit{exo} in the \textit{actuators velocity} control mode.
\end{itemize}
The trials lasted 3 minutes to ensure subjects had reached steady-state walking; data were collected during the last minute. 
Before data acquisition, subjects familiarized with the exoskeleton and practised the task in all the conditions. 
For each condition, they decided the preferred walking speed on the treadmill. 
For the \textit{exo} conditions, they tried different values of the gain $K_{myo}$ that tunes the assistance and selected the one that meets their desired support.

\subsection{Metrics}
To test the hypotheses and evaluate the exoskeleton control strategy in assisting carrying, we access the users' lumbar muscle activity and gait kinematics in the different assistance modes. 
Moreover, subjective perceptions are investigated with an ad-hoc questionnaire to highlight any discomfort caused by the exoskeleton assistance.

The activity of the lumbar muscles (right and left erector spinae longissimus lumborum (LL)) is measured with surface EMG electrodes (BTS FREEEMG, BTS Bioengineering, Italy), attached following SENIAM guidelines \cite{Stegeman2007_ReferenceSENIAM}.
EMG data processing includes band-pass filtering (10-400 Hz), filtering to remove the electrical noise and the electrocardiography (ECG) signal \cite{Drake2006_ECGelimination}, rectification and low-pass filtering (2.5 Hz cut-off frequency \cite{Potvin1996_EMGAnalysis}). 
Finally, EMG signals are normalized to the maximum voluntary contraction (MVC) \cite{McGill1991_MVCtechniquesTrunk}. 
MVC was acquired prior to data collection during a maximum exertion task repeated three times: subjects lay in a prone position with the torso hanging over the edge of a test bench and extended their trunk upward against manual resistance \cite{McGill1991_MVCtechniquesTrunk}. 
Muscle activities during the different assistive modes are compared in terms of root mean square (RMS) and 90th percentile value computed within one minute.
These metrics provide valuable measures of the signal amplitude and are reliable indicators of muscle involvement during physically demanding tasks \cite{Ranavolo2018_EMGandLiftingIndex}.
The RMS estimates the average muscle activity, while the 90th percentile captures the activity related to the maximal effort. It is preferred to the maximum value because it is more robust to outliers. 

Motion data are measured with Xsens device, a 3D motion tracking system (Xsens, The Netherlands).
Subjects wear 8 Xsens Inertial Measurement Units (IMUs), which record the position and orientation of the feet, shanks, thighs, pelvis, and trunk. 
Using a biomechanical model of the subject, the Xsens software reconstructs lower limb kinematics and gait events (i.e., foot contacts with the ground).
The kinematics analysis assesses the average hip and knee RoM and stride length with comparison between the \textit{no-exo} and \textit{exo} modes.

The questionnaire, filled-in by the subjects at the end of each \textit{exo} mode, is a simplified version of the RPE (Rate of Perceived Exertion) form, as used in \cite{Poliero2020_XoTrunkCarrying}.
It investigates the users' perceptions of benefit or hindrance caused by the exoskeleton assistance at different body regions: shoulders, arms, back/trunk, and legs. 

The statistical analyses are conducted using SPSS 20.0 (IBM SPSS) with significance threshold set at p $<$0.05. 
Firstly, the Shapiro-Wilk normality test evaluates whether the data were normally distributed. 
Since the data are normally distributed, for each investigated parameter (stride length, hip and knee RoM, speed, and RMS and 90th percentile of the erector spinae longissimus activity), a one-way repeated-measures analysis of variance (ANOVA) is performed to determine if there is any significant effect of the exoskeleton assistance mode on each of these parameters. 
Where the ANOVA test shows a main effect, the post hoc analysis is carried out with Bonferroni’s correction.

\section{RESULTS}
In this Section, we present the results of the experimental testing.
Some of the metrics analyzed are the same used in the previous study by Poliero et al. \cite{Poliero2020_XoTrunkCarrying} in order to compare the outcomes.
The study by Poliero et al. \cite{Poliero2020_XoTrunkCarrying} assisted carrying task with the same exoskeleton but using a different control i.e., symmetric torque assistance.

\label{sec:results}
\subsection{Gait kinematics}
The kinematics analysis highlights whether there was a significant change in the gait movement due to the exoskeleton assistance.
The average changes in stride length between \textit{no-exo} and the two \textit{exo} conditions ($<$1\%) did not reveal any main effect of the exoskeleton assistance (F\textsubscript{(2,18)} = 1.111, p = 0.351).
Likewise, hip RoM was not significantly changed by the \textit{exo} factor (F\textsubscript{(2,18)} = 3.036, p = 0.073).
Conversely, for the knee RoM the one-way repeated measures ANOVA showed an effect of the \textit{exo} factor (F\textsubscript{(1.099,9,890)} = 22.464, p = 0.001). 
The post hoc analysis revealed significant reductions up to 8\% for the knee RoM with both the \textit{foot sensors} and \textit{actuators velocity} modes (both p = 0.001). 
As regard the preferred task speed, only one subject decided a lower speed for the task with the exoskeleton, while 9 subjects walked at the same speed in the \textit{no-exo} and \textit{exo} conditions (F\textsubscript{(2,18)} = 1.000, p = 0.387).
Results are summarized in Table \ref{table_kinematics}.
For the three assistance conditions, the means (and standard deviations) of stride length, hip and knee RoMs and task execution speed are presented. 
The F and p-values are the results of the repeated measures ANOVA tests with the assistance condition as the independent variable.

\begin{table}[h]
\fontsize{6.5pt}{6.5pt}\selectfont
\caption{Mean (std) of stride length, hip and knee roms and speed}
\vspace{-13pt}
\label{table_kinematics}
\begin{center}
\begin{tabular}{|c|c|c|c|c|c|}
\hline
 & \textit{no-exo} & \textit{foot sensors} & \textit{actuators v.} & F & p-value\\ 
\hline
s. length [m] & 1.07 (0.11) & 1.06 (0.12) & 1.08 (0.14) & 1.111 & 0.351 \\ 
hip RoM [\textdegree]& 32.11 (3.37) & 31.47 (4.05) & 33.96 (3.78) & 3.036 & 0.073\\ 
knee RoM [\textdegree]& 59.76 (6.69) & 54.95 (6.77) & 55.70 (6.40) & 22.464 & \textbf{0.001}\\ 
speed [m/s] & 0.74 (0.14) & 0.73 (0.13) & 0.73 (0.13)& 1.000 & 0.387 \\ 
\hline
\end{tabular}
\end{center}
\end{table}

\subsection{Subjective perception}
\begin{figure}[thpb]
	\centering
        \includegraphics[scale=0.4]{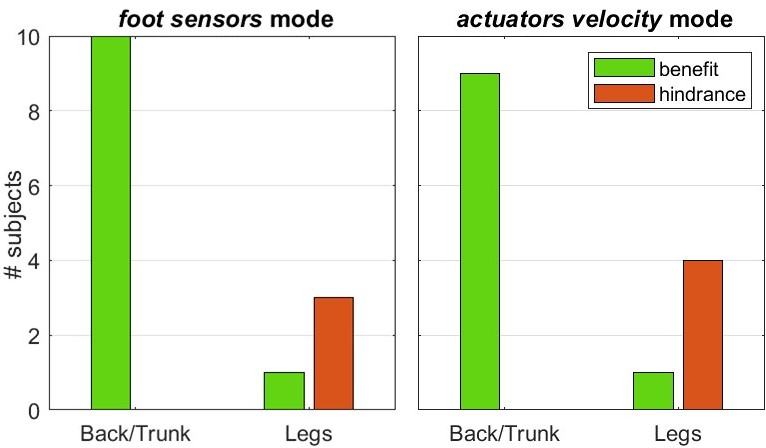}
	\caption{\label{fig_subjectivePerception} Users' perception questionnaire results with \textit{exo} assistance for \textit{foot sensors} and \textit{actuators velocity} modes. Each body region is divided in those who felt comfort (green) and those who felt hindrance (red).}
\end{figure}

Fig. \ref{fig_subjectivePerception} displays for back/trunk and legs how many subjects reported a benefit or a hindrance when using the exoskeleton. 
All the subjects felt a benefit and no hindrance on the back with both \textit{exo} modes.  
On the legs, hindrance was perceived by 3 and 4 subjects with the \textit{foot sensors} and \textit{actuators velocity} modes, respectively, while one subject reported a benefit.

\subsection{Muscle activity}
\begin{figure}[thpb]
	\centering
        \includegraphics[scale=0.5]{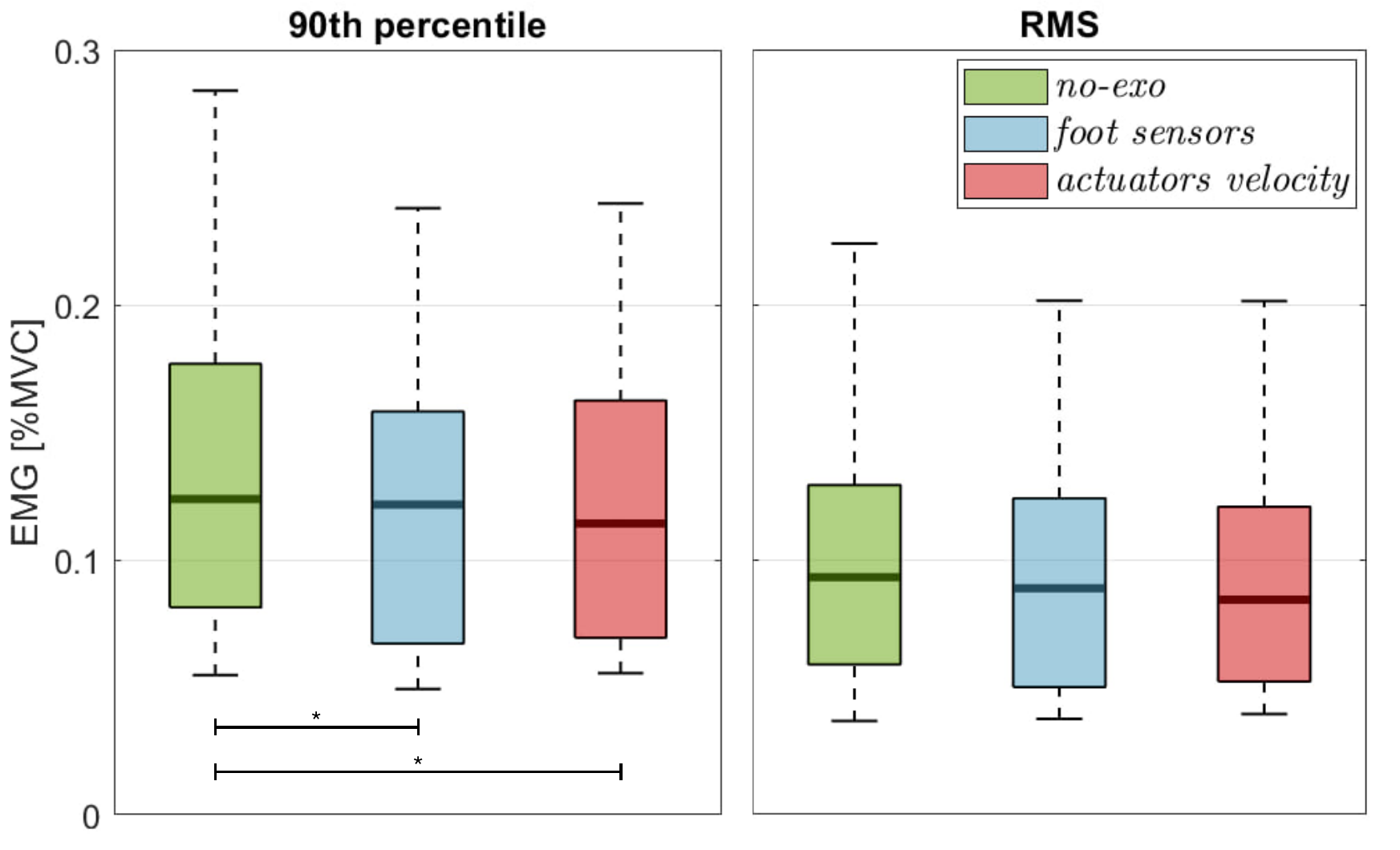}
	\caption{\label{fig_EMGresults} Box plots of the muscle activity metrics, RMS and 90th percentile, in the \textit{no-exo} (green), \textit{foot sensors} mode (blue), and \textit{actuators velocity} mode (red). For each box, the central line represents the sample median, the edges of the box are the 25th and 75th percentiles, and the whiskers are the min and the max values. Horizontal bars indicate the level of statistical significance (* = p-value $<$0.05).}
\end{figure}

The RMS and 90th percentile EMG across all the subjects are shown in Fig. \ref{fig_EMGresults}.
The one-way repeated measures ANOVA showed an effect of the \textit{exo} factor for the 90th percentile muscle activity (F\textsubscript{(2,18)} = 5.316, p = 0.015).
The post hoc analysis showed significant reductions for the peak muscle activity (90th percentile) equal to 11\% and 10\% for the \textit{foot sensors} and \textit{actuators velocity} modes, respectively, compared to the \textit{no-exo} assistance (p = 0.010 and p = 0.028, respectively). 
In summary, out of 10, 9 and 8 subjects (respectively for the \textit{foot sensors} and \textit{actuators velocity} modes) had reduced muscle activity due to exoskeleton assistance. 
A similar trend was registered for the RMS values, which estimate the average muscle activity within the trial, however it was not statistically significant (F\textsubscript{(2,18)} = 2.793, p = 0.088).


\section{DISCUSSION}
\label{sec:Discussion}
With respect to state-of-the-art, significant improvements were obtained in assisting the carrying task with a back-support exoskeleton. 
In particular, analyzing the kinematics of the gait, the exoskeleton only slightly conditioned the natural movement.
Differently from previous studies \cite{Baltrusch2019_LAEVOEvaluationLiftingWalking} \cite{Poliero2020_XoTrunkCarrying}, stride length, hip RoM and speed did not change significantly between \textit{no-exo} and \textit{exo} assistance modes.
Moreover, the significant reductions found for the knee RoM were lower than those obtained when a symmetric torque control strategy was used to assist carrying (8 versus 12\%) \cite{Poliero2020_XoTrunkCarrying}.

Likewise, subjective perception results indicate that the users' hindrance during walking was substantially reduced: only 30 and 40\% of the participants reported hindrance and discomfort in the legs when walking with the exoskeleton.
This percentage was equal to 78\% in the prior study \cite{Poliero2020_XoTrunkCarrying}, which interviewed 9 subjects with the same RPE questionnaire.
In the study of Baltrusch et al. \cite{Baltrusch2018_LaevoEffect}, evaluating the commercial Laevo device, walking with the exoskeleton had significantly increased general discomfort and perceived difficulty and reduced objective performance.

The reduction of the peak EMG activity of the erector spinae muscle with the exoskeleton was statistically significant when compared to the \textit{no-exo} condition for both \textit{exo} modes. 
By contrast, the reduction of the mean muscle activity was not significant but showed a similar trend (with p = 0.088).
The comparison with the subjective data confirms this beneficial effect of the exoskeleton. 
Indeed, all the subjects reported to have felt benefit on the back compared to the execution of the task without the exoskeleton aid.

In summary, the outcomes of the experimental analysis for the control strategy using the two different frameworks were similar.
The assistance provided by both the \textit{exo} modes only slightly conditioned the users' gait kinematics, received good results in terms of subjective perception, and significantly reduced the participants' lumbar load.
Indeed, the two \textit{exo} modes modulate the torque in the same manner, i.e., with (\ref{eq:tauRightLeft}) and (\ref{eq:tauTotal}).
The difference is that they are based on different gait phase detection frameworks.
These gait phase detection frameworks have different limitations and advantages.
The main limitation of the \textit{foot sensors} framework is that it requires force sensors to be inserted into the user's shoes.
However, its main advantage is that the gait segmentation is always reliable with this type of sensor.
Conversely, the main advantage of the \textit{actuators velocity} framework is that it does not require any additional sensor that the user must wear. The velocity of the actuators is measured with encoders embedded in the exoskeleton, making it particularly convenient for application in real working environments.
However, while gait phase detection is accurate and reliable for treadmill walking, its performance must be validated for less structured movements; in real working scenarios, carrying tasks are performed in sequence with other activities, at different speeds or intermittently.   

The limitations of this work are due to the novelty of assisting carrying task that presents different challenges in the context of back-support exoskeletons. 
Differently from symmetric tasks, e.g. lifting, while there is no doubt that the back is the body region more overloaded, the asymmetry of the legs' movement makes it difficult to provide the assistance to the lumbosacral joint without interfering or hindering in some way the legs, where the exoskeleton is anchored.
To minimize discontinuity in the assistance, we decided to provide to each leg during single stance the same assistive torque as in double stance, even if the load on the leg in single stance is double.
Still, because of leg phase transitions, the assistive torques are discontinuous on the legs, which may be perceived as uncomfortable or awkward.
A possible solution to these discontinuities could be to turn on and off slowly the assistance from 0 to $\tau_{stance}$ and from $\tau_{stance}$ to 0, introducing, however, a certain delay in both providing full assistance for stance and removing assistance to facilitate swing.
Furthermore, back-support exoskeletons are expected to perform better with tasks that involve trunk bending. As the effort required from the back muscles is mainly due to balancing the moment generated by gravity (acting on both the user's upper body and on the eventual external object), muscle activation can be easily and significantly reduced with gravity-compensation assistance.
In this context, a larger sample size would help obtain more robust and reliable results.
Moreover, users with more experience with the exoskeleton or with the task itself would be more aware of the best way to use the device (e.g., by selecting a more appropriate gain for assisting the task).


\section{CONCLUSIONS}
\label{sec:conclusions}
Repetitive and demanding MMH activities, performed daily in different industrial sectors, significantly load workers' musculoskeletal systems and in particular the back, increasing the risk of developing chronic LBP.
In industry 4.0, back-support exoskeletons represent a high-potential solution to the prevention and mitigation of LBP. However, the applicability of back-support exoskeletons to activities differing from lifting is still an open issue.

This work takes the first step towards the use of a back-support exoskeleton for assisting carrying tasks.
We implemented a new control strategy on an active back-support exoskeleton. 
The strategy modulates the assistance based on two different gait phase detection frameworks. 
One framework uses \textit{foot sensors} as input to perform gait phase detection; the other framework uses the exoskeleton \textit{actuators velocity}.
Using these frameworks allows to assist the user's legs according to their gait phases. 
Differently from previous works, this control strategy assists the user's back with asymmetrical forces on the legs.
The new strategy has been assessed with an experimental evaluation. 
The presented results show that the exoskeleton assistance decreased the load on the users’ back, with minimal hindrance for the legs.

Interestingly, one gait phase detection framework is based on signals (i.e., the actuators velocity) from sensors embedded in the exoskeleton and thus the related control does not require any additional sensor that must be worn by the user.
This advantage makes it particularly convenient for a real working scenario. 
Where full automation is unfeasible or prohibitively expensive, smart factories can exploit collaborative exoskeletons to promote safety, efficiency, and productivity.

Future works will test the strategy more extensively. 
To meet the requirements of real industrial scenarios, the gait phase detection framework based on \textit{foot sensors} may be simplified by reducing the number of force sensors. 
Indeed, the relevant information for exoskeleton control is whether the foot is in contact with the ground; therefore, precise data on foot forces are unnecessary.
A possible simplified improvement could be using only two sensors, one on the front of the foot and one on the back, to capture heel strike and toe-off.
On the other hand, the framework based on \textit{actuators velocity} should be tested for more complex execution of carrying task. 
In this context, a possible improvement could be obtained using machine learning algorithms, integrating the exoskeleton actuators' velocity with other signals from embedded sensors (e.g., IMUs). 




\bibliographystyle{IEEEtran}
\bibliography{IEEEabrv, References.bib}


\end{document}